\documentclass{article} % For LaTeX2e
\usepackage{iclr2019_conference}
\usepackage{times}
\iclrfinalcopy
\usepackage{tabularx}
\usepackage{adjustbox}
\usepackage{xparse}
\usepackage{amsmath}
\usepackage{amsthm}
\usepackage{amssymb}
\usepackage{xspace}
\usepackage{relsize}
\usepackage{graphicx}
\usepackage{multirow}
\usepackage{url}
\usepackage{comment}
\usepackage{amsmath,amsfonts,bm}

\def\eqref#1{equation~\ref{#1}}
\def\1{\bm{1}}

\def\rv{{\textnormal{v}}}

\def\rvv{{\mathbf{v}}}

\def\ervv{{\textnormal{v}}}

\def\rmV{{\mathbf{V}}}

\def\vo{{\bm{o}}}

\def\vv{{\bm{v}}}

\def\vx{{\bm{x}}}
\def\vy{{\bm{y}}}

\def\evx{{x}}

\def\mV{{\bm{V}}}

\def\mX{{\bm{X}}}
\def\mY{{\bm{Y}}}

\DeclareMathAlphabet{\mathsfit}{\encodingdefault}{\sfdefault}{m}{sl}
\SetMathAlphabet{\mathsfit}{bold}{\encodingdefault}{\sfdefault}{bx}{n}

\newcommand{\pdata}{p_{\rm{data}}}
\newcommand{\pmodel}{p_{\rm{model}}}

\newcommand{\E}{\mathbb{E}}

\newcommand{\R}{\mathbb{R}}

\DeclareMathOperator*{\argmin}{arg\,min}

\usepackage[draft]{minted}
\usepackage{xr}

\newcommand{\subsubparagraph}[1]{}

\makeatletter
\let\@myref\ref

\newcommand{\refsec}[1]{Sec.\,\@myref{#1}}
\newcommand{\refseq}[1]{Sec.\,\@myref{#1}}
\newcommand{\refig}[1]{Fig.\,\@myref{#1}}
\newcommand{\refigs}[2]{Fig.\,\@myref{#1}-\@myref{#2}}

\newcommand{\reftbl}[1]{Table~\@myref{#1}}
\newcommand{\refstep}[1]{Step~\@myref{#1}}
\newcommand{\refalgo}[1]{Algorithm~\@myref{#1}}
\newcommand{\refchap}[1]{Chapter~\@myref{#1}}
\newcommand{\reflst}[1]{List~\@myref{#1}}
\newcommand{\refeq}[1]{Eq.\@myref{#1}}

\makeatother

\newcounter{list}[section]

\newcommand{\brackets}[1]{{\left<#1\right>}}
\newcommand{\braces}[1]{{\left\{#1\right\}}}
\newcommand{\parens}[1]{{\left(#1\right)}}

\newcommand{\defun}[1]{%
\makeatletter
\expandafter\def\csname the#1\endcsname{\text{\it #1}}
\expandafter\def\csname #1\endcsname ##1{\csname the#1\endcsname\left(##1\right)}%
\makeatother
}

\newcommand{\defsetop}[2]{%
\makeatletter% variable, set, descriptor
 \expandafter\def\csname #1\endcsname ##1##2##3{%
  \expandafter\def\csname #1arg\endcsname{##1}%
  \expandafter\def\csname #1set\endcsname{##2}%
  \expandafter\def\csname #1cond\endcsname{##3}%
  \braces{##1##2\mid #2 ##3}%
 }%
\makeatother%
}

\defun{precond}
\defun{condition}
\defun{params}
\defun{type}
\defun{proc}
\defun{cost}

\defun{push}
\defun{pref}
\defun{init}
\defun{goal}
\defun{objects}
\defun{task}

\defun{parent}
\defun{plateau}

\defsetop{filter}{}
\defsetop{map}{}

\newcommand{\pddl}[1]{\textsf{\small #1}}

\def\_{\\[-0.3em]}

\newcommand{\ipc}[1][2011]{IPC#1\xspace}

\makeatletter

\newcommand{\newheuristic}[2]{%
 \def#1{%
  \ifmmode%
  h^\text{#2}\xspace%
  \else%
  \text{#2}\xspace%
  \fi%
 }%
}

\newheuristic{\lmcut}{LMcut}
\newheuristic{\mands}{M\&S}
\newheuristic{\pdb}{PDB}
\newheuristic{\ff}{FF}
\newheuristic{\ce}{CEA}
\newheuristic{\cg}{CG}
\newheuristic{\ad}{add}
\newheuristic{\lc}{LC}

\newcommand{\newUnitCostHeuristic}[2]{%
 \def#1{%
  \ifmmode%
  \hat{h}^\text{#2}\xspace%
  \else%
  \text{#2}\xspace%
  \fi%
 }%
}

\newUnitCostHeuristic{\lmcuto}{LMcut}
\newUnitCostHeuristic{\mandso}{M\&S}
\newUnitCostHeuristic{\ffo}{FF}
\newUnitCostHeuristic{\ceo}{CEA}
\newUnitCostHeuristic{\cgo}{CG}
\newUnitCostHeuristic{\ado}{add}
\newUnitCostHeuristic{\gco}{GoalCount}
\newUnitCostHeuristic{\lco}{LC}

\makeatother

\def\B{\textbf}

\def\mathdef{\overset{\text{def}}{=}}

\def\ref{\todo{Do not use ``ref'' directly!}}
\author{Masataro Asai \\ IBM Research}
\title{Set Cross Entropy: Likelihood-based Permutation Invariant Loss Function for Probability Distributions}

\begin{document}
\maketitle

%%%%%%%%%%%%%%%%%%%%%%%%%%%%%%%%%%%%%%%%%%%%%%%%%%%%%%%%%%%%%%%%
\begin{comment}

* Maintain 80 characters / line.
 
* too much ``''s make the sentence look scattered and visually less recognizable. ``e.g.'' also.

* \em, \bf, \it are all obsolete \TeX primitives, and it does not take effect properly --- for example, {\bf {\it aaa}} shows ``aaa'' in italic but NOT IN BOLD. Use \emph{}, \textit{}, \textbf{} and so on.

* always use \ff, \fd, \cea, \pr, \mv , and do not use it directly, e.g. FF, FD/LAMA2011, etc. 

* use of footnotes should be minimized.

* IPC2011 should always be \ipc . The definition can later be modified in abbrev.sty .

* prefer separated words over hyphened words. domain
  independent>domain-independent, planner independent >
  planner-independent.

* Table, Figure, Fig., should not be used directly. Always use \refig and \reftbl. When the development flag is enabled, direct use of \ref signals an error.

* Caption ends with a period.

ICLR note:

    The recommended paper length is 8 pages, with unlimited additional pages for citations.

    There will be a strict upper limit of 10 pages.

    Reviewers will be instructed to apply a higher standard to papers in excess of 8 pages.

    Authors may use as many pages of appendices (after the bibliography)
    as they wish, but reviewers are not required to read these.

\end{comment}

%%%%%%%%%%%%%%%%%%%%%%%%%%%%%%%%%%%%%%%%%%%%%%%%%%%%%%%%%%%%%%%%

\begin{abstract}
We propose a permutation-invariant loss function designed for the neural
networks reconstructing
% a set of elements as
a set of elements without considering the order within its vector representation. Unlike popular
approaches for encoding and decoding a set, our work does not
rely on a carefully engineered network topology nor
by any additional sequential algorithm.
 % other than the stochastic gradient descent for optimization.
 % 
The proposed method, Set Cross Entropy, has a natural information-theoretic interpretation
and is related to the metrics defined for sets.
 % 
% Instead, we show that a simple loss
% function which fits in one line and can be used with a normal decoder 
% is sufficient.
 % 
% We also propose an enhancement to the runtime complexity of the proposed loss function
% from $O(n^2)$ to $O(n\log n)$ with an appropriate preprocessing methods such as spatial trees or hierarchical clustering
% in order to increase the scalability of the proposed approach to the larger sets.
We evaluate the proposed approach in two object reconstruction tasks and a rule learning task.
\end{abstract}

\section{Introduction}

Sets are fundamental mathematical objects which appear frequently in the real-world dataset.
However, there are only a handful of studies on learning a set representation in the machine learning literature.
% In particular, while existing work focuses on handling the set input,
% to the best of our knowledge,
% none of the published work is targeting the permutation-invariant set reconstruction.
% 
% 
% 
In this study, we propose a new objective function called \emph{Set Cross Entropy} (SCE)
to address the permutation invariant set generation.
SCE measures the cross entropy between two sets that consists of multiple
elements, where each element is represented as a multi-dimensional
probability distribution in $[0,1] \subset \R$ (a closed set of reals between 0,1).
SCE is invariant to the object permutation, therefore does not distinguish
two vector representations of a set with the different ordering.
The SCE is simple enough to fit in one line
and can be naturally interpreted as
a formulation of the log-likelihood maximization between two sets
derived from a logical statement.

% \begin{figure}[b]
%  \centering
%  \includegraphics[width=0.5\linewidth]{img/concept.pdf}
%  \caption{A loss function that does not consider the permutation invariance makes the training harder.}
% \end{figure}

The SCE loss trains a neural network in a permutation-invariant manner, and the network learns to output a set.
Importantly, this is \emph{not} to say that the neural network \emph{learns to represent}
a function that is permutation-invariant with regard to the input.
The key difference in our approach is that we allow the network to output a vector representation of a set
that may have a different ordering than the examples used during the training.
In contrast, previous studies focus on
learning a function that returns the same output for the different permutations of the input elements.
Such scenarios assume that
an output value at some index is matched against the target value at the same index.
\footnote{
For example, the permutation-equivariant/invariant layers in \citet{zaheer2017deep}
are evaluated on the classification tasks, the regression tasks and the set expansion tasks.
In the classification and the regression tasks, the network predicts a single value, which has no ordering.
In the set expansion tasks, the network predicts the probability $p_i$ for each tag $i$ in the output vector.
Thus, reordering the output does not make sense.
}

This characteristic is crucial in the tasks where the objects included
in the supervised signals (training examples for the output) do not have
any meaningful ordering.
For example, in the logic rule learning tasks,
a first-order logic horn clause
does not care about the ordering inside the rule body since
logical conjunctions are invariant to permutations,
e.g.
$\pddl{father}(\pddl{c},\pddl{f})\leftarrow \parens{\pddl{parent}(\pddl{c},\pddl{f}) \land \pddl{male}(\pddl{f})}$
and
$\pddl{father}(\pddl{c},\pddl{f})\leftarrow \parens{\pddl{male}(\pddl{f}) \land \pddl{parent}(\pddl{c},\pddl{f})}$
are equivalent.

% Another scenario that such characteristic becomes important is 
% when the set of elements may be reported in a different order in each observation.
% 
% An example of such a scenario may be
% even if the 
% 
% objects in the environment
% such as the 

To apply our approach,
no special engineering of the network topology is required
other than the standard hyperparameter tuning.
The only requirement is that the target output examples are
the probability vectors in $[0,1]^{N\times F}$, which is easily addressed
by an appropriate feature engineering including autoencoders with softmax or sigmoid latent activation.

We demonstrate the effectiveness of our approach in
two object-set reconstruction tasks and
the supervised theory learning tasks that learn to perform the backward chaining of the horn clauses.
% a variety of set reconstruction tasks
% including object set reconstruction, Theory learning tasks and the variational modeling of 
% the sparse representation of the 3D point cloud.
In particular, we show that the SCE objective is superior to the training using
the other set distance metrics,
including Hausdorff and set average (Chamfer) distances.

% In this study, we claim that a simple modified loss function that is
% invariant to the object ordering is sufficient for generating a set of
% elements.  The new objective function called \emph{Set Cross Entropy}
% measures the cross entropy between two sets that consists of multiple
% elements where each element is represented as a multi-dimensional
% probability distribution.

\section{Backgrounds and Related Work}

\subsection{Learning a Set Representation}

Previous studies try to discover the appropriate structure for the neural networks
that can represent a set.
Notable recent work includes permutation-equivariant / invariant layers that
addresses the permutation in the input \citep{guttenberg2016permutation,ravanbakhsh2016deep,zaheer2017deep}.

% A function $f(X)$ is \emph{permutation equivariant} when $X,f(X)\in \mathcal{D}^n$ for some $n$
% and $\forall \pi; \pi(f(X))=f(\pi(X))$.
Let $X$ be a vector representation of a set $\braces{x_1,\ldots, x_n}$
and $\pi$ be an arbitrary permutation function for a sequence.
A function $f(X)$ is \emph{permutation invariant} when $\forall \pi; f(X)=f(\pi(X))$.
\citet{zaheer2017deep} showed that functions are permutation-invariant iff it can be decomposed into a form
\[
 f(X) = \rho \parens{\sum_{x\in X} \phi(x)}
\]
where $\rho, \phi$ are the appropriate mapping function.
% and all permutation-equivariant functions into a form
% \[
%  f(X) = (\lambda \bm{I} + \gamma \bm{1} {\bm{1}}^T) X \qquad \lambda,\gamma\in\R, \bm{1}=[1,1,\ldots, 1]^T \in \R^n
% \]

However, as mentioned in the introduction,
the aim of these layers is to learn the functions that are permutation-invariant with regard to the input permutation,
and not to reconstruct a set in a permutation-invariant manner (i.e. ignoring the ordering).
In other words,
permutation-equivariant/invariant layers are only capable of encoding a set.

\citet{probst2018set} recently proposed a method dubbed as ``Set Autoencoder''.
It additionally learns a permutation matrix that is applied before the output
so that the output matches the target.
The target for the permutation matrix is generated by a Gale-Shapley greedy stable matching algorithm,
which requires $O(n^2)$ runtime.
The output is compared against the training example
with a conventional loss function such as binary cross entropy or mean squared error,
which requires the final output to have the same ordering as the target.
Therefore, this work tries to learn the set as well as the ordering between the elements,
which is conceptually different from learning to reconstruct a set while ignoring the ordering.

% Sequential algorithms like Gale-Shapley requires $O(n^2)$ runtime.
% or special sequential algorithms that explicitly tries to learn the permutation,
% such as permutation matrices produced by Gale-Shapley algorithm \citep{probst2018set}

Another line of related work utilizes
Sinkhorn iterations \citep{adams2011ranking,santa2017deeppermnet,mena2018learning}
in order to directly learn the permutations.
Again, these work assumes that the output is generated in a specific order (e.g. a sorting task),
which does not align with the concept of the set reconstruction.

% 17cvpr DeepPermNet: Visual Permutation Learning
% the purpose is to learn the permutation and reconstruct the correct ordering,
% not to ignore the permutation.
% 
% 18iclr gumbel-sinkhorn
% the purpose is to learn the permutation and reconstruct the correct ordering,
% not to ignore the permutation.

\subsection{Set Distance}

Set distances / metrics are the binary functions that satisfy the metric axioms.
They have been utilized for measuring the visual object matching or for feature selection
\citep{huttenlocher1993comparing,dubuisson1994modified,piramuthu1999hausdorff}.
Note that, however, in this work, we use the informal usage of the terms ``distance'' or ``metric''
for any non-negative binary functions that may not satisfy the metric axioms.

% \begin{align}
%  \textit{Non-Negativity}:\quad      & d(x,y)\geq 0.                 \\
%  \textit{Identity}:\quad            & x=y \Leftrightarrow d(x,y)=0. \\
%  \textit{Symmetry}:\quad            & d(x,y) = d(y,x).              \\
%  \textit{Triangle inequality}:\quad & d(x,y)+d(y,z) \geq d(x,z).
% \end{align}

There are several variants of set distances.
Hausdorff distance between sets \citep{huttenlocher1993comparing} is a function that satisfies the metric
axiom.  For two sets $X$ and $Y$, the directed Hausdorff distance with an element-wise distance $d(x,y)$
is defined as follows:
\[
 \mathcal{H}_{1d}(X,Y) = \max_{x\in X} \min_{y\in Y} d(x,y)
\]
% The undirected Hausdorff distance is then defined as:
% \[
%  H_d(X,Y) = \max \braces{H_{1d}(X,Y), H_{1d}(Y,X)}.
% \]
The element-wise distance $d$ is Euclidean distance or Hamming distance, for example,
depending on the target domain.
% When $d$ satisfies the metric axiom, the Hausdorff
% distance as a whole also satisfies the metric axiom.

Set average (pseudo) distance \citep[Eq.(6)]{dubuisson1994modified},
also known as Chamfer distance, is
a modification of the original Hausdorff
distance which aggregates the element-wise distances by summation.
The directed version is defined as follows:
\[
 A_{1d}(X,Y) = \frac{1}{|X|}\sum_{x\in X} \min_{y\in Y} d(x,y).
\]
% , while several more complex extensions are
% also available, and also the undirected version is defined similar to the Hausdorff distance.
% While set distances were shown to work well on certain applications,
% to our knowledge, they were not applied in the context of neural networks and set reconstruction.
% As we later show in the experiments, they do not work well in this scenario.
% In particular, the maximization part of the Hausdorff distance causes the same effect as using the max-norm
% as the loss function for neural networks, which causes the entire value of the network
Set average distance has been used for image matching,
as well as to autoencode the 3D point clouds in the euclidean space
for shape matching \citep{zhu2016deep}.

% Earth Mover distance \citep{rubner2000earth} is a distance defined on 
% a set of bins, originally designed for image matching
% to reflect the characteristics of the human visual perception.
% It can be seen as measuring the cost of the solution of a transportation problem,
% the task of moving the minimum amount of soil between the different set of bins.
% The total cost is defined as the sum of the work required to move the soil,
% which is a product of the amount of the soil moved and the distance moved.
% Therefore, Earth Mover distance is a natural choice when
% the amount of soil and the distance between the bins can be meaningfully defined,
% such as the pixel values and the Euclid distance between the pixels for image matching,
% or similarly for the voxel values for 3D object matching.
% Since this is not the case with the general probability distribution,
% we do not elaborate further on this distance.

\section{Set Cross Entropy}

\label{sce}

Inspired by the various set distances, we 
propose a straightforward formulation of likelihood maximization between two sets of probability distributions.
In what follows, we define the cross entropy between two sets $\mX,\mY\in [0,1]^{N\times F}$, where
$[0,1]$ is a closed set of reals between 0 and 1.

Let $\mathcal{X}=\braces{\mX^{(1)},\mX^{(2)},\ldots}$ be the training dataset,
and $\mY$ be the output matrix of a neural network.
Assume that each $\mX\in \mathcal{X}$ consists of $N$ elements where each element is
represented by $F$ features, i.e. $\mX=\braces{\vx_1 \ldots \vx_N}, \vx_i\in
\mathbb{R}^F$.  We further assume that $\vx_i\in [0,1]^F$ by a suitable transformation,
e.g., feature learning using an autoencoder with the sigmoid activation added to the latent layer.
The set $\mX$ actually takes the vector representation, which essentially makes $\mX\in [0,1]^{N\times F}$.
In this paper, we focus on the Bernoulli distribution, where each feature $\evx_{i,f}$ represents the probability that
the corresponding binary random variable $\ervv_{i,f}$ being true, i.e. $\evx_{i,f}=P(\ervv_{i,f}=1)$, which also means
$P(\ervv_{i,f}=0)=1-\evx_{i,f}$. However,
the proposed method naturally extends to the multinomial case
where each elementary feature could be represented as a probability vector that sums to 1
instead of a single value in $[0,1]$.

For simplicity, we assume that the number of elements in the set $\mX$ and $\mY$ is known and fixed to $N$.
Therefore $\mY$ is also a matrix in $[0,1]^{N\times F}$.
Furthermore, we assume that $\mX$ is preprocessed and contains no duplicated elements.
 
In practice, if $|\mX|$ varies across the dataset,
it suffices to take $N^{\max}=\max_{\mX\in\mathcal{X}} |\mX|$,
the largest number of elements in $\mX$ across the dataset $\mathcal{X}$,
and add the dummy, distinct elements $d_0\ldots d_{N^{\max}}$ to fill in the blanks.
For example, when there are $N$ objects of $F$ features and we want to normalize the size of the set to $M (> N)$,
one way is to add an additional axis to the feature vector ($F+1$ features) where the additional
$F+1$-th feature is 0 for the real data and 1 for the dummy data, and the additional $M-N$ objects are generated
in an arbitrary way (e.g. as a binary sequence 100000, 100001, 100010, 100011, ... for $F=5$).

% subsection{Similarity between Vectors}

Let $\rvv \in \braces{0,1}^F$ a $F$-dimensional binary random variable,
where $\pdata(\rv_i=1) = x_i$, $\pmodel(\rv_i=1) = y_i$, and $\vx,\vy\in[0,1]^F$.
For measuring the similarity between two $F$-dimensional probability distributions $\pdata$ and $\pmodel$,
the natural loss function would be the cross entropy or the negative log likelihood,
denoted as $\mathrm{H}(\vx,\vy)$.
\begin{align}
\begin{split}
 \mathrm{H}(\vx,\vy)
 &= \E_{\rvv\sim\pdata}\brackets{-\log \pmodel (\rvv=\vv)} \\
 &= \E_{\rvv\sim\pdata}\brackets{-\log \pmodel \parens{\bigwedge_{i=1}^F \rv_i=v_i}} 
 % &= \E_{\rvv\sim\pdata}\brackets{-\sum_{i=1}^F \log \pmodel \parens{\rv_i=v_i}} \\
 % &= \sum_{v \in \braces{0,1}} \sum_{i=1}^F - \pdata(\rv_i=v) \log \pmodel(\rv_i=v)\\
 = \sum_{i=1}^F - x_i \log y_i - (1-x_i) \log (1-y_i).
\end{split}
\end{align}

However, applying it directly to the matrices $\mX,\mY\in[0,1]^{N\times F}$ by flattening them into vectors
(e.g. $\text{flatten}(\mX) \in [0,1]^{N\cdot F}$) results in an undesired outcome.
The cross entropy loss has only a single global optima because it does not consider the permutations between $N$ objects,
e.g., for $\mX=[\vo_1,\vo_2,\vo_3]$ ($\vo_i\in [0,1]^F$),
$\mY=[\vo_2,\vo_3,\vo_1]$ is not the global minima of $\mathrm{H}(\text{flatten}(\mX),\text{flatten}(\mY))$.
Previous approach \citep{probst2018set} tried to solve this problem by learning an additional permutation matrix
that ``fixes'' the order, basically requiring to memorize the ordering.

We take a different approach of directly fixing this loss function.
The target objective is to maximize the probability of two sets being equal, thus ideally,
at the global minima, two sets should be equal.
The key to understand our main contribution in the next section is that the above expansion
implicitly assumes a certain definition of the equivalence between vectors:
\[
 \rvv = \vv \quad \Longleftrightarrow \quad \bigwedge_{i=1}^F \rv_i=v_i
\]
We argue that, in order to measure the similarity (e.g. likelihood,
cross entropy) between sets, we similarly need to start from
the very definition of the equivalence between sets.

% subsection{Similarity between Sets}

% \input{unused/t-s-normal.tex}
Equivalence of two sets $S,T$ is defined as:
\[
 S=T \quad\Leftrightarrow\quad  \parens{{S\subseteq T}             \ \land\  {S\supseteq T}} 
     \quad\Leftrightarrow\quad  \parens{\forall s \in S ; s \in T} \ \land \ \parens{\forall t \in T ; t \in S}.
\]
However, under the assumption that $|S|=|T|=N$ and $T$ contains $N$ distinct elements (no duplicates),
$S\supseteq T$ is a sufficient condition for $S=T$. (Proof: If $S\supseteq T$ and $S\not\subseteq T$,
there are some $s'\in S$ such that $s'\not\in T$. Since $N$ distinct elements in $T$ are also included in $S$, $s'$ becomes $S$'s
$N+1$-th element, which contradicts $|S|=N$. Note that this proof did not depend on the distinctness of $S$'s elements.)
Under this condition, therefore,
\begin{align}
\begin{split}
 S=T \quad \Leftrightarrow   \quad S\supseteq T
     \quad \Leftrightarrow & \quad \forall t \in T ;\; t \in S               \\
     \quad \Leftrightarrow & \quad \forall t \in T ;\; \exists s \in S ;\; s=t \\
     \quad \Leftrightarrow & \quad \bigwedge_{t\in T}\; \bigvee_{s\in S}\; s=t .
\end{split}
\end{align}

We now translate this logical formula into the corresponding log likelihood.
Assume a $N\times F$ dimensional, binary random variable $\rmV=[\rvv_1, \ldots \rvv_N]$,
and its value $\mV = [\vv_1 \ldots \vv_N] \in \braces{0,1}^{N\times F}$.
The target value we aim to compute is the following negative log likelihood
\begin{align}
 \E_{\rmV\sim \pdata}\brackets{-\log \pmodel(\rmV=\mV)}
 &= \sum_{\mV \in \braces{0,1}^{N\times F}} -\pdata(\rmV=\mV) \log \pmodel(\rmV=\mV).
\label{eq:setce-base}
\end{align}
Since $\pdata$ produces the dataset $\mathcal{X}$, it satisfies $\pdata(\rmV=\mV)=0$ when $\mV$ contains duplicates.
This allows us to ignore those $\mV$s from the summation and also perform the following transformation:
\begin{align}
\log \pmodel(\rmV=\mV)
&= \log \pmodel(\rmV\supseteq\mV)
 =\log \pmodel(\bigwedge_{i=1}^N          \bigvee_{j=1}^N            \rvv_j=\vv_i)\nonumber\\
&=            \sum_{i=1}^N   \log \pmodel(\bigvee_{j=1}^N            \rvv_j=\vv_i)\nonumber\\
&\geq         \sum_{i=1}^N   \log      \sum_{j=1}^N          \pmodel(\rvv_j=\vv_i)\nonumber\\
&=            \sum_{i=1}^N   \log      \sum_{j=1}^N \exp\log \pmodel(\rvv_j=\vv_i)\nonumber\\
&=            \sum_{i=1}^N \text{logsumexp}_{j=1}^N     \log \pmodel(\rvv_j=\vv_i).
\label{eq:setce-inequality}
\end{align}
The inequality comes from ignoring the possibility that two random vectors taking the same value.
The equality holds at the global minima.
Next we compute the expectation $\E_{\rmV\sim \pdata}$ of this sum:
\begin{align}
 \E_{\rmV\sim \pdata}\brackets{-\log \pmodel(\rmV=\mV)}
 &\leq \E_{\rmV\sim \pdata} \brackets{-\sum_i \text{logsumexp}_j     \log \pmodel(\rvv_j=\vv_i)} \quad \because \text{\refeq{eq:setce-base}, \refeq{eq:setce-inequality}}\\
 &= - \sum_i \E_{\rmV  \sim \pdata}   \brackets{\text{logsumexp}_j     \log \pmodel(\rvv_j=\vv_i)}\\
 % &= - \sum_i \sum_{\mV\in\braces{0,1}^{N\times F}} \pdata(\rmV=\mV) \text{logsumexp}_j     \log \pmodel(\rvv_j=\vv_i)\\
 &= - \sum_i \E_{\rvv_i\sim \pdata} \brackets{\text{logsumexp}_j     \log \pmodel(\rvv_j=\vv_i)}\\
 &\leq - \sum_i \text{logsumexp}_j \E_{\rvv_i\sim \pdata} \brackets{\log \pmodel(\rvv_j=\vv_i)} \\
 &= -\sum_i \text{logsumexp}_j \parens{-H(\vx_i,\vy_j)}
 % &= -\sum_i \text{logsumexp}_j \parens{\sum_{f=1}^F x_{j,f}\log y_{i,f} + (1-x_{j,f}) \log (1-y_{i,f})}\\
 \mathdef \mathrm{SH}(\mX,\mY). \label{eq:setce}
\end{align}
The inequality is due to Jensen's inequality applied to logsumexp, which is convex.
This gives the upper bound of the negative log likelihood between sets
% , which matches at the global minima.
which we call the \emph{Set Cross Entropy}.

% subsection{Analyzing Set Cross Entropy}

Set Cross Entropy has the following characteristics:
First, compared to the original cross entropy loss, which has the single global minima,
SCE has exponentially many number of global minima
by making every permutations of the global minima also the global minima.

Next, notice that logsumexp is a smooth upper approximation of the maximum,
therefore the set average distance $A_{1\mathrm{H}}$ is an upper bound of $\mathrm{SH}$\footnote{The equality holds only for $N=1$.}:
\begin{align}
       - \sum_i \text{logsumexp}_j \parens{-H(\vx_i,\vy_j)}
 &\leq - \sum_i \max_j \parens{-H(\vx_i,\vy_j)} 
  =      \sum_i \min_j \parens{ H(\vx_i,\vy_j)}
\nonumber\\
\therefore
 \mathrm{SH}(\mX,\mY) &\leq N\cdot A_{1\mathrm{H}}(\mX,\mY).
\label{eq:setavgce}
\end{align}
Intuitively, this is because
the set average returns a value which does not account for the
possibility that the current closest $\vy=\argmin_\vy \mathrm{H}(\vx,\vy)$ of $\vx$ may not
converge to the $\vx$ in the future during the training.

%% This example does not work.
% We illustrate this by comparing two examples:
% Let
% $\mX=\braces{[0],[0]}$, $\mY_1=\braces{[0.1],[0.1]}$ and $\mY_2=\braces{[0.1],[0.9]}$.
% % ce(x1,y11) -0log0.1 - 1log0.9 = -log0.9 = 0.105360516 = log 1/0.9
% % ce(x1,y12) -0log0.1 - 1log0.9 = -log0.9
% % logsumexp (-ce) = log (0.9+0.9) = log 1.8 = 0.587786665
% % ce(x2,y11) -0log0.1 - 1log0.9 = -log0.9 = 0.105360516 = log 1/0.9
% % ce(x2,y12) -0log0.1 - 1log0.9 = -log0.9
% % logsumexp = log1.8
% % -sum logsumexp -ce = -2log1.8 = -1.17557333 % ??? WTF ???
% %                         -------- this is because it violates the assumption that \mX do not contain duplicates

We illustrate this by comparing two examples:
Let
$\mX=\braces{[0,1],[0,0]}$, $\mY_1=\braces{[0.1,0.5],[0.1,0.5]}$ and $\mY_2=\braces{[0.1,0.5],[0.9,0.5]}$.
The set cross entropy (\refeq{eq:setce}) reports the smaller loss for $\mathrm{SH}(\mX,\mY_1)=-\log 0.81\approx 0.09$ than for $\mathrm{SH}(\mX,\mY_2)=-\log0.25\approx 0.60$.
This is reasonable because the global minima is given when the first axis of both $y$s are 0 ---
$\mY_2$ should be more penalized than $\mY_1$ for the $0.9$ in the second element.
In contrast, $A_{1\mathrm{H}}(\mX,\mY)$ considers only the closest element ($\argmin_{\vy\in \mY}\mathrm{H}(\vx,\vy) = [0.1,0.5]$) for each $\vx$,
therefore returns the same loss $=-\log 0.2025\approx0.69$ for both cases, ignoring $[0.9,0.5]$ completely.
In fact, $A_{1\mathrm{H}}(\mX,\mY)$ has zero gradient at $\mY=\braces{[0,0.5],[y,0.5]}$ for any $y\in[0,1]$.
% although this is an unstable equilibrium and would rarely happen in practice due to the initial randomization.

Furthermore, the following inequality suggests that the traditional cross entropy between the matrices $\mX$ and $\mY$
% which does not consider the object permutation
is an upper bound of the set average of the cross entropies,
therefore is an even looser upper bound of the set cross entropy. Here, $\vx_i,\vy_i$
are the $i$-th element of the vector representation of $\mX$ and $\mY$,
respectively:
\begin{align*}
 \mathrm{SH}(\mX,\mY)
 &\leq \sum_i \min_j \mathrm{H}(\vx_i,\vy_j) &\because \text{\refeq{eq:setavgce}}\\
 &\leq \sum_i \mathrm{H}(\vx_i,\vy_i)        &\because \forall i; \ \min_j \mathrm{H}(\vx_i,\vy_j) \leq \mathrm{H}(\vx_i,\vy_i)
\end{align*}
This gives a natural interpretation that ignoring the permutation reduces the cross entropy.

Finally, the directed Hausdorff distance based on cross entropy ($\mathcal{H}_{1H}$)
is also an upper bound of SCE and the set average:
\begin{align*}
 A_{1\mathrm{H}}(\mX,\mY) = \frac{1}{N} \sum_i \min_j H(\vx_i,\vy_j) \leq \max_i \min_j H(\vx_i,\vy_j) = \mathcal{H}_{1H}(\mX,\mY).
\end{align*}

\section{Evaluation}

\subsection{Object Set Reconstruction}

The purpose of the task is to obtain the latent representation of a set of objects and reconstruct them,
where each object is represented as a feature vector.
We prepared two datasets originating from classical AI domains: 
Sliding tile puzzle (8-puzzle) and Blocksworld.

Learning to reason about the object-based, set representation of the environment
is crucial in the robotic systems
that continuously receive the list of visible objects from the visual perception module (e.g. \citet[YOLO]{redmon2016you}).
In a real-world systems,
appropriate handling of the set is necessary because
it is unnatural to assume that the objects in the environments are always reported in the same order.
In particular, the objects even in the \emph{same} environment state may be reported
in various orders if multiple such modules are running in parallel in an asynchronous manner.

% is gaining popularity in the reinforcement learning community,
% which adds a structure representing the relations between the objects.
% % 
% For example,
% Deep Symbolic RL \cite{garnelo2016towards} showed that a hand-crafted ``common sense prior'' (e.g. proximity) accelerates DRL.
% Relation Networks \cite{santoro2017simple} combine two elements in the feature maps of a convolutional neural network \cite{lecun2015deep} and output real values.
% Deep Relational RL \cite{zambaldi2018relational} combines Relation Networks and attention-based message passing.
% % Relational Inductive Bias \cite{battaglia2018relational} shows that DRL is enhanced by a hand-crafted, explicit graph representation of the input and the Graph Neural Networks \cite{scarselli2009graph}.
% % Another line of work in incorporating relations into machine learning models is its application in common-sense physics modeling.
% Relational Neural Expectation Maximization \cite{van2018relational} enumerates pairs of objects
% to model the relations in the environment for common-sense physics modeling.

In this experiment, we show that the permutation invariant loss function like SCE is necessary
for learning to reconstruct a set in such a scenario.
In this setting, a network is required to reconstruct a set from a single latent representation,
while the objects as the target output
may be randomly reordered each time the same set is observed and presented to the neural network.

\subsubsection{8 Puzzle}

Each feature vector as an object consists of 15 features, 9 of which represent the tile number (object ID) and the remaining 6
represent the coordinates.
Each data point has 9 such vectors, corresponding to the 9 objects in a single tile configuration.
The entire state space of the puzzle is 362880 states.
We generated 5000 states and used the 4500 states as the training set.

\begin{figure}[htbp]
 \centering
 \includegraphics[width=0.7\linewidth]{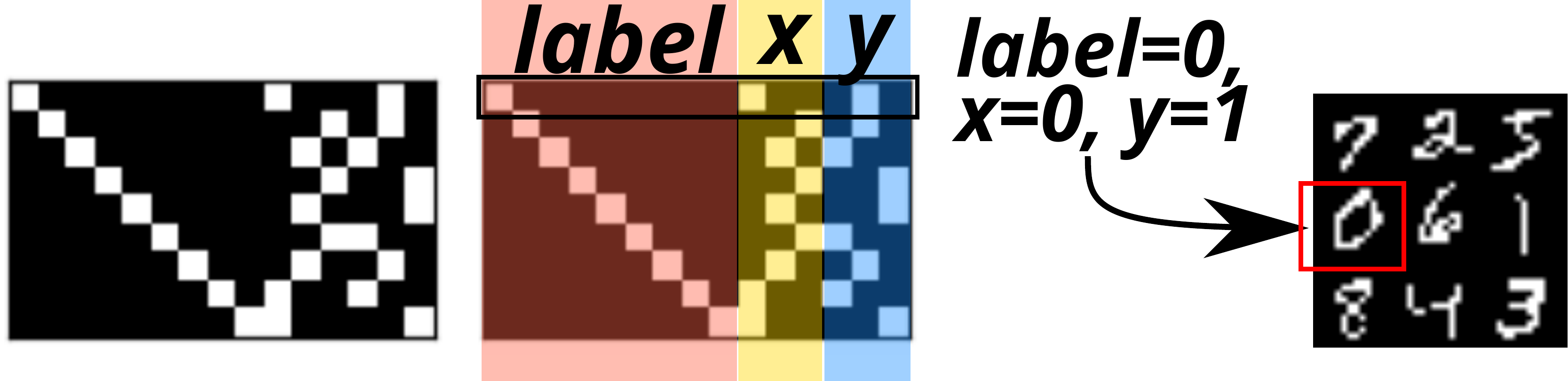}
 \caption{
A single 8-puzzle state as a 9x15 matrix, representing 9 objects of 15 features.
The first 9 features are the tile numbers
and the other 6 features are the 1-hot x/y-coordinates.
}
 \label{8puzzle-features}
\end{figure}

We prepared
% a full connected autoencoder and
an autoencoder with
the permutation invariant layers \citep{zaheer2017deep} as the encoder and
the fully-connected layers as the decoder.
Since it uses a permutation-invariant encoder,
the latent space is already guaranteed to learn a representation
that is invariant to the input ordering.
The key question here is then whether they can be robustly trained against
the random permutations in the training examples for the output.

We tested the reconstruction ability in four scenarios:
\textbf{(1)} In the first scenario, the dataset is provided in a standard manner.
\textbf{(2)} In the second scenario, 
we augment the input dataset by repeating the elements 5 times
and randomly reorder the object vectors in each set.
The randomized dataset is used as the input to the network, while the target output is still the original dataset (repeated 5 times, without reordering).
The purpose of this experiment is to verify the claim of the Deep Set \citep{zaheer2017deep}
that it is able to handle the input in a permutation invariant manner.
In order to compensate the datasize difference, the maximum training epoch is reduced by 1/5 times compared to the first scenario.
\textbf{(3)} In the third scenario, we apply the similar operation to the target output of the network.
Essentially we always feed the input in the same fixed order while forcing it to learn from the randomized target output.
Each time the same data is presented, the target output has the different ordering while the input has the fixed ordering.
Therefore, the training should be performed in such a way that the ordering in the output is properly ignored.
\textbf{(4)} Finally, in the fourth scenario, the ordering in both the input and the output are randomized.

We trained the same network with four different loss functions,
\textbf{(a)} the traditional cross entropy $\mathrm{H}$,
\textbf{(b)} Set Cross Entropy $\mathrm{SH}$,
\textbf{(c)} directed set average of the cross entropy $A_{1\mathrm{H}}$ and
\textbf{(d)} the directed Hausdorff measure of the cross entropy $\mathcal{H}_{1\mathrm{H}}$,
resulting in 16 training scenarios in total.
We performed the same experiment 10 times and took the statistics.
The purpose of this is to address the potential concern about the stability of the training.
We kept the same set of training/testing data, and the only difference between the runs is the random seed.

We measured the rate of the successful reconstruction among the entire dataset in the above 16 scenarios.
The ``successful reconstruction'' is defined as follows:
Recall that every data point is a discrete binary vector in the 8-Puzzle dataset while
the output of the network is a continuous $N\times F$ matrix of reals between 0 and 1.
Therefore, we round the output of the network to $0/1$ and directly compare the result with the input.
If every object vector in a set is matched by some of the output object vector, then it is counted as a success.

The training with the standard cross entropy loss ($\mathrm{H}$) succeeds in cases \textbf{(1,2)} while failed in cases \textbf{(3,4)}.
The case \textbf{(2)} reproduces the claim in \citep{zaheer2017deep} that it encodes the input in an permutation-invariant manner,
while it failed in the latter cases because the training is not permutation-invariant with regard to the output.
In contrast, the training with the Set Cross Entropy loss succeeds in all cases.
This shows that the permutation-invariant loss function is necessary for training a network with a dataset consisting of sets.
In this dataset, the training with set average distance $A_{1\mathrm{H}}$ also succeeded
because it is a looser but still effective upper-bound of the negative log likelihood.
% However, in one of the 10 runs, $A_{1\mathrm{H}}$ did not converge,
% showing that the A1H (baseline) could be unstable, possibly due to the issue
% explained in the example at the end of section 3. % , though this is a speculation from the empirical result.
In contrast,
the training with Hausdorff distance failed to learn the representation at all.
% The reason is suspected that the maximization in Hausdorff distance reduced
% the training speed.

\begin{table}[htbp]
 \centering
 \relsize{-0.5}
   \begin{tabular}{|c|cccc|}
    \hline
Target ordering
  & \multicolumn{2}{c}{Fixed} & \multicolumn{2}{c|}{Random} 
 \\\hline
Input ordering
  & \textbf{(1)}Fixed & \textbf{(2)}Random 
  & \textbf{(3)}Fixed & \textbf{(4)}Random 
 \\\hline
\textbf{(a)}$\mathrm{H}$                & \B{1.00} & \B{1.00} & 0.00      & 0.00     \\
\textbf{(b)}$\mathrm{SH}$               & \B{1.00} & \B{1.00} & \B{1.00}  & \B{1.00} \\
\textbf{(c)}$A_{1\mathrm{H}}$           & \B{1.00} & \B{1.00} & \B{1.00}  & \B{1.00} \\
\textbf{(d)}$\mathcal{H}_{1\mathrm{H}}$ & 0.00     & 0.00     & 0.00      & 0.00     \\
\hline
  \end{tabular}
 \caption{The best reconstruction success ratio in the 10 runs of 16 training scenarios.
Among the four loss functions, best results are shown in \B{bold}.
$\mathrm{SH}$ and $A_{1\mathrm{H}}$ both succeeded to reconstruct the binary vectors in the 8 puzzles.
The traditional cross entropy $\mathrm{H}$ and the Hausdorff distance $\mathcal{H}_{1\mathrm{H}}$ 
both failed to reconstruct the binary vectors.
}
\label{8puzzle-success}
\end{table}

Next, to address the claim that the network is able to learn from the dataset with the variable set size,
we performed an experiment which applies the dummy-vector scheme (\refsec{sce}).
In this experiment, we modified the dataset to model such a scenario by randomly dropping one to five elements out of 9 elements.
The maximum number of elements is 9.
The dropping scheme is specified as follows:
Out of the 5000 states generated in total (including the training / testing dataset),
approximately half of the states have 9 tiles, $1/4$ of the states have 8 tiles, ... and $1/2^5$ of the states have 5 tiles.
The elements to drop are selected randomly.

The results in \reftbl{8puzzle-variable} shows that the training with our proposed $\mathrm{SH}$ loss function
achieves the best success ratio for the reconstruction when the output order is randomized.
The reconstruction includes the dummy vectors, indicating that the
network is able to represent not only the elements in the set but also the number of the missing elements.

\begin{table}[htbp]
 \centering 
 \relsize{-0.5}
   \begin{tabular}{|c|cccc|}
    \hline
Target ordering
  & \multicolumn{2}{c}{Fixed} & \multicolumn{2}{c|}{Random} 
 \\\hline
Input ordering
  & \textbf{(1)}Fixed & \textbf{(2)}Random 
  & \textbf{(3)}Fixed & \textbf{(4)}Random 
 \\\hline
\textbf{(a)}$\mathrm{H}$                & \B{0.76} & \B{0.85} & 0.12      & 0.00     \\
\textbf{(b)}$\mathrm{SH}$               & 0.60     & 0.65     & \B{0.62}  & \B{0.62} \\
\textbf{(c)}$A_{1\mathrm{H}}$           & 0.57     & 0.56     & 0.58      & 0.56     \\
\textbf{(d)}$\mathcal{H}_{1\mathrm{H}}$ & 0.00     & 0.00     & 0.00      & 0.00     \\
\hline
  \end{tabular}
 \caption{The best reconstruction success ratio in the 10 runs of 16 training scenarios, where the size of the set randomly varies from 4 to 9 in the dataset.
Among the four loss functions, best results are shown in \B{bold}.
The proposed $\mathrm{SH}$ achieved the best success rate overall,
$A_{1\mathrm{H}}$ comes next, 
the traditional cross entropy $\mathrm{H}$ and the Hausdorff distance $\mathcal{H}_{1\mathrm{H}}$ both failed in most cases.
}
\label{8puzzle-variable}
\end{table}

\subsubsection{Blocksworld}

\begin{figure}[htbp]
 \centering
 \includegraphics[width=0.7\linewidth]{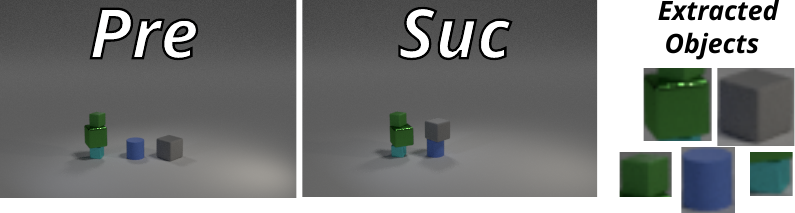}
 \caption{
An example Blocksworld transition.
Each state has a perturbation from the jitter in the light positions and the ray-tracing noise.
Objects have the different sizes, colors, shapes and surface materials.
Regions corresponding to each object in the environment are extracted
according to the bounding box information included in the dataset
generator output, but is ideally automatically extracted by object
recognition methods such as YOLO \citep{redmon2016you}.  Other objects
may intrude the extracted regions.
} \label{blocks-example}
\end{figure}

In order to test the reconstruction ability for the more complex
feature vectors, we prepared a photo-realistic Blocksworld dataset
(\refig{blocks-example})
% which was generated with a code modified from the CLEVR dataset generator
% \cite{johnson2017clevr} 
which contains the blocks world states rendered by Blender 3D engine.
There are several cylinders or cubes of
various colors and sizes and two surface materials (Metal/Rubber)
stacked on the floor, just like in the usual STRIPS \citep{McDermott00} Blocksworld domain.
In this domain, three actions are performed:
\texttt{move} a block onto another stack or on the floor,
and \texttt{polish}/\texttt{unpolish} a block i.e. change the
surface of a block from Metal to Rubber or vice versa.
All actions are applicable only when the block is on top of a stack or on the floor.
The latter actions allow changes in the non-coordinate features of the object vectors.

The dataset generator produces a 300x200 RGB image and a state description which contains
the bounding boxes (bbox) of the objects.
Extracting these bboxes is a object recognition task we do not address in this paper,
and ideally, should be performed by a system like YOLO \citep{redmon2016you}.
We resized the extracted image patches in the bboxes to 32x32 RGB,
compressed it into a feature vector of 1024 dimensions with a convolutional autoencoder, then
concatenated it with the bbox $(x_1,y_1,x_2,y_2)$
which is discretized by 5 pixels and encoded as 1-hot vectors (60/40 categories for $x$/$y$-axes),
resulting in 1224 features per object.
The generator is able to enumerate all possible states
(80640 states for 5 blocks and 3 stacks).
We used 2250 states as the training set and 250 states as the test set.

\begin{figure}[htbp]
 \centering
\textbf{(a)} \includegraphics[width=0.2\linewidth]{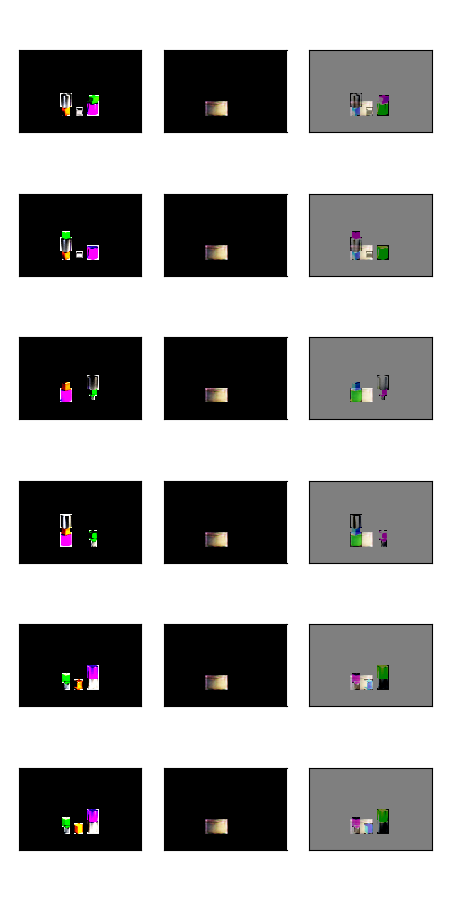}
\textbf{(b)} \includegraphics[width=0.2\linewidth]{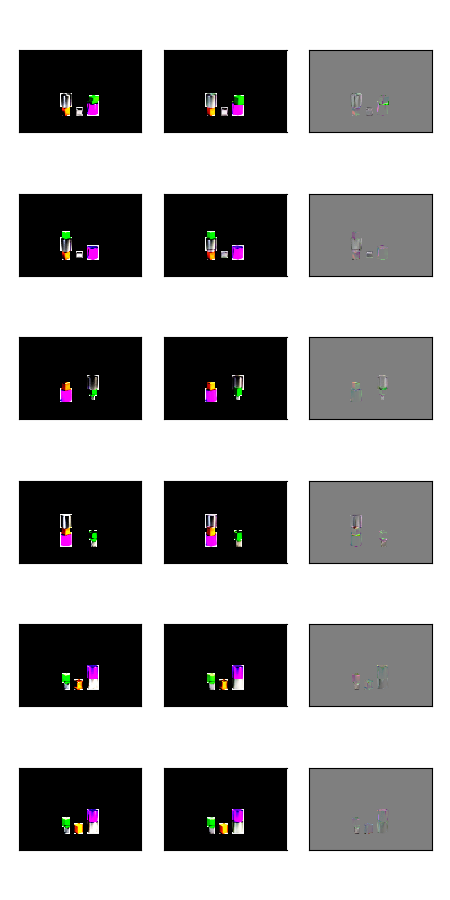}
\textbf{(c)} \includegraphics[width=0.2\linewidth]{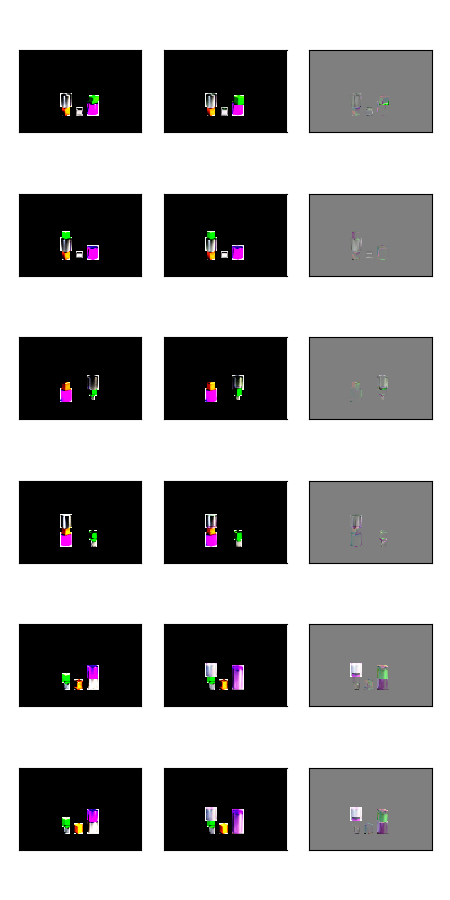}
\textbf{(d)} \includegraphics[width=0.2\linewidth]{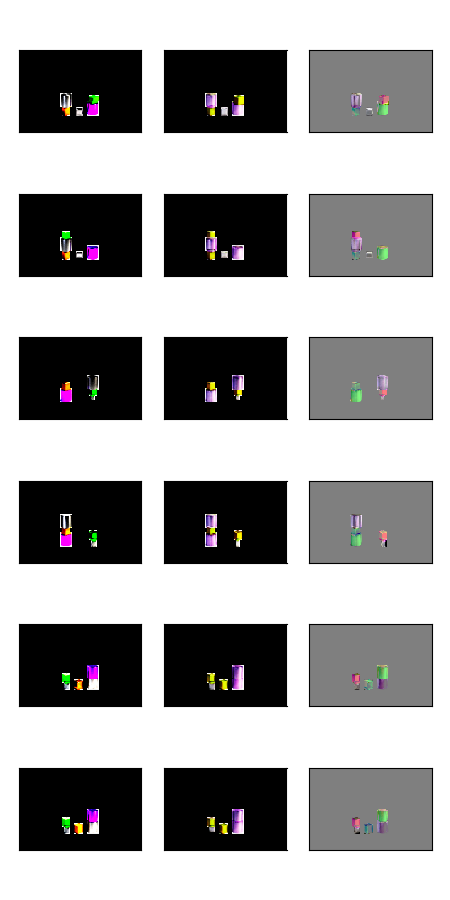}\\ 
 \caption{
The visualizations of the Blocksworld state input (left), its reconstruction (middle) and their pixel-wise difference (right).
From the left, each three columns represent 
\textbf{(a)} the traditional cross entropy $\mathrm{H}$,
\textbf{(b)} Set Cross Entropy $\mathrm{SH}$,
\textbf{(c)} directed set average of the cross entropy $A_{1\mathrm{H}}$ and
\textbf{(d)} the directed Hausdorff measure of the cross entropy $\mathcal{H}_{1\mathrm{H}}$.
The proposed \textbf{(b)} Set Cross Entropy correctly reconstructs the input.
}
 \label{blocks-reconstruction}
\end{figure}

We also verified the results qualitatively.
Some reconstruction results are visualized in \refig{blocks-reconstruction}.
These visualizations are generated by
pasting the image patches decoded from the first 1024 axes of the reconstructed 1224-D feature vectors
in a position specified by the reconstructed bounding box in the last 200 axes.

In \reftbl{blocks-visual-reconstruction-error},
we measured the difference between the visualization results of the input
and the output (as shown in \refig{blocks-reconstruction}) by the Root Mean
Squared Error of the pixel values averaged over RGB, pixels and the dataset.
Each pixel is represented in the $[0,1]$ range
(closed set of reals between 0 and 1), thus the 0.1 on the table means that 
pixels differ by 0.1 on average.

\begin{table}[htbp]
 \centering
 \relsize{-0.5}
 \begin{tabular}{|c|cccc|}
  \hline
Target ordering
  & \multicolumn{2}{c}{Fixed} & \multicolumn{2}{c|}{Random} \\\hline
Input ordering
  & \textbf{(1)}Fixed & \textbf{(2)}Random 
  & \textbf{(3)}Fixed & \textbf{(4)}Random 
 \\\hline
\textbf{(a)}$\mathrm{H}$                & 0.10     & 0.10     & 0.15     & 0.15     \\
\textbf{(b)}$\mathrm{SH}$               & \B{0.08} & \B{0.08} & \B{0.07} & \B{0.08} \\
\textbf{(c)}$A_{1\mathrm{H}}$           & \B{0.08} & 0.10     & \B{0.08} & \B{0.08} \\
\textbf{(d)}$\mathcal{H}_{1\mathrm{H}}$ & 0.14     & 0.15     & 0.16     & 0.15     \\
\hline
 \end{tabular}
 \caption{RMSE between the visualized images, the best results of 10 runs.
Both $\mathrm{SH}$ and
$A_{1\mathrm{H}}$ successfully converged below the sufficient accuracy.
}
\label{blocks-visual-reconstruction-error}
\end{table}

\subsection{Rule Learning ILP Tasks}

\defun{isa}

The purpose of this task is to learn to generate the prerequisites (body) of the first-order-logic horn clauses
from the head of the clause.
Unlike the previous tasks, this task is not an autoencoding task.
The bodies are considered as a set because the order of the terms inside a body does not matter
for the clause to be satisfied.

The main purpose of this experiment is to show the effectiveness of our approach on set generation,
not to demonstrate a more general neural theorem proving system.
An interesting avenue of future work is to see how our approach can help the existing work
on neural theorem proving.

We used a Countries dataset \citep{bouchard2015approximate} that contains 163 countries
and trained the models for $n$-hop neighbor relations.
For example, for $n=2$,
given a head
$\pddl{neighbor2}(\pddl{austria},\pddl{germany},\pddl{belgium})$ as an input,
the task is to predict the body
$\braces{\pddl{neighborOf}(\pddl{austria},\pddl{germany}),\pddl{neighborOf}(\pddl{germany},\pddl{belgium})}$,
which is a set of two terms.
This is a weaker form of a more general backward chaining used in Neural Theorem Proving \citep{rocktaschel2017end}
because the output does not contain free variables.
% However, existing methods in Neural Theorem Proving are currently limited to the rules that contain just one or two terms in the body.

In the $n$-neighbor scenario,
the input is a $2+163(n+1)$-dimensional vector,
which consists of
a one-hot label of 2 categories for the predicate of the head,
and $n+1$ one-hot labels of 163 categories for the arguments of the head.
% 163 categories corresponds to the number of countries in the Countries dataset.
For example,
a head $\pddl{neighbor2}(\pddl{austria},\pddl{germany},\pddl{belgium})$ spends 2 dimensions for identifying the predicate
\pddl{neighbor2}, and three 1-hot vectors of 163 categories for representing \pddl{austria},\pddl{germany},\pddl{belgium}.
The output is a $n \times 328$ matrix, where each row represents a binary predicate $(328 = 2 + 2\cdot 163)$ such as
$\braces{\pddl{neighborOf}(\pddl{austria},\pddl{germany}),\pddl{neighborOf}(\pddl{germany},\pddl{belgium})}$ for $n=2$.
Each element uses 2 dimensions for identifying the predicate head \pddl{neighborOf} and
two 1-hot vectors of 163 categories for the arguments (e.g. \pddl{austria} and \pddl{germany}).

% For simplicity, we assume that a head has only a single body that consists of only conjunctions,
% unlike traditional Prolog programs that allow multiple rules for a single head (an equivalent program can be written with disjunctions, which is not allowed either).
%  % 
% All terms that appear in a body are binary predicates and are represented as 3-hot vectors.
% Each vector is a concatenation of 1-hot encoding of the predicate or the 2-hot encoding of the arguments.
% % Similar to the blocksworld task, we create a compressed representation for each term with an autoencoder.
% % All predicates are unary or binary, where for the unary predicates the unused second argument points to the \texttt{nil} symbol.
% The namespaces for the predicates and the arguments are separated.

We trained the network with the \pddl{neighbor}-$n$ datasets ranging from $n=2$ to $n=5$ (see the result table for the detailed domain characteristics).
During the inference, the index of the largest output of the softmax is treated as the answer.
We counted the ratio of the clauses across the test set where every body term matches against one of the output terms.
The output data (body terms) may have an arbitrary ordering, and we have another variant similar to the previous experiment:
In the randomized body order dataset, the dataset is repeated 5 times, while the ordering of the terms inside each body
is randomly shuffled.

\reftbl{countries-reconstruction-error} shows that the network with Set Cross Entropy
achieved the best accuracy, set average generally comes in the second and other losses struggled.
This trend was observed both in the the randomized and the fixed-ordering dataset.
This shows that the Set Cross Entropy relaxes the search space by
adding more global minima and making the training easier.

\newcolumntype{Y}{>{\centering\arraybackslash}X}

\begin{table}[htbp]
 \centering
 \relsize{-0.5}
 % \begin{adjustbox}{width={\linewidth},keepaspectratio}
   \begin{tabularx}{\linewidth}{|c|Y|Y|Y|Y|}
    \hline
  & \multicolumn{2}{c|}{$n=2$, $\pddl{neighbor2}(a,b,c)\pddl{:-}\ldots $}      & \multicolumn{2}{c|}{$n=3$, $\pddl{neighbor3}(a,b,c,d)\pddl{:-}\ldots$}    
 \\                                                                            
  & \multicolumn{2}{c|}{Dataset: 2858 ground clauses}                          & \multicolumn{2}{c|}{Dataset: 11000 ground clauses}
 \\
  & \multicolumn{2}{c|}{Training: 2250 clauses; Test: 250 clauses. }           & \multicolumn{2}{c|}{Training: 2250 clauses; Test: 250 clauses. }
 \\
  & \multicolumn{1}{c|}{Fixed  }
  & \multicolumn{1}{c|}{Random }
  & \multicolumn{1}{c|}{Fixed  }
  & \multicolumn{1}{c|}{Random }
 \\
\textbf{(a)}$\mathrm{H}$                &0.32     & 0.36                                   &0.10     & 0.10     \\                           
\textbf{(b)}$\mathrm{SH}$               &\B{0.96} & \B{0.94}                               &\B{0.72} & \B{0.70} \\                           
\textbf{(c)}$A_{1\mathrm{H}}$           &0.91     & \B{0.94}                               &0.61     & 0.60     \\                           
\textbf{(d)}$\mathcal{H}_{1\mathrm{H}}$ &0.87     & 0.85                                   &0.55     & 0.57     \\                           
\hline
  & \multicolumn{2}{c|}{$n=4$, $\pddl{neighbor4}(a,b,c,d,e)\pddl{:-}\ldots$}   & \multicolumn{2}{c|}{$n=4$, $\pddl{neighbor4}(a,b,c,d,e)\pddl{:-}\ldots$}
 \\
  & \multicolumn{2}{c|}{Dataset: 39878 ground clauses}                         & \multicolumn{2}{c|}{Dataset: 39878 ground clauses}
 \\                    
  & \multicolumn{2}{c|}{Training: 2250 clauses; Test: 250 clauses. }           & \multicolumn{2}{c|}{Training: 9000 clauses; Test: 1000 clauses. }       
 \\
  & \multicolumn{1}{c|}{Fixed  }
  & \multicolumn{1}{c|}{Random }
  & \multicolumn{1}{c|}{Fixed  }
  & \multicolumn{1}{c|}{Random }
 \\
\textbf{(a)}$\mathrm{H}$                &0.02     & 0.03      &0.04     & 0.04     \\                         
\textbf{(b)}$\mathrm{SH}$               &\B{0.38} & \B{0.36}  &\B{0.87} & \B{0.86} \\                         
\textbf{(c)}$A_{1\mathrm{H}}$           &0.33     & 0.34      &0.81     & 0.79     \\                         
\textbf{(d)}$\mathcal{H}_{1\mathrm{H}}$ &0.22     & 0.24      &0.50     & 0.53     \\                         
\hline
  & \multicolumn{2}{c|}{$n=5$, $\pddl{neighbor5}(a,b,c,d,e,f)\pddl{:-}\ldots$}   & \multicolumn{2}{c|}{$n=5$, $\pddl{neighbor5}(a,b,c,d,e,f)\pddl{:-}\ldots$}
 \\
  & \multicolumn{2}{c|}{Dataset: 137738 ground clauses}                          & \multicolumn{2}{c|}{Dataset: 137738 ground clauses}
 \\                     
  & \multicolumn{2}{c|}{Training: 2250 clauses; Test: 250 clauses. }             & \multicolumn{2}{c|}{Training: 9000 clauses; Test: 1000 clauses. }         
 \\
  & \multicolumn{1}{c|}{Fixed  }
  & \multicolumn{1}{c|}{Random }
  & \multicolumn{1}{c|}{Fixed  }
  & \multicolumn{1}{c|}{Random }
 \\
\textbf{(a)}$\mathrm{H}$                &0.00     & 0.01      &0.01     & 0.01     \\                           
\textbf{(b)}$\mathrm{SH}$               &\B{0.17} & \B{0.15}  &\B{0.65} & \B{0.60} \\                           
\textbf{(c)}$A_{1\mathrm{H}}$           &0.13     & 0.13      &0.53     & 0.56     \\                           
\textbf{(d)}$\mathcal{H}_{1\mathrm{H}}$ &0.06     & 0.06      &0.20     & 0.18     \\                           
\hline
  \end{tabularx}
 % \end{adjustbox}
 \caption{
The rate of correct answers on the test set, best results of 10 runs.
``Random'' indicates that the target output is shuffled.}
\label{countries-reconstruction-error}
\end{table}

\section{Discussion}

\citet{vinyals2016order} repeatedly emphasized the advantage of
limiting the possible equivalence classes of the outputs
by engineering the training data for solving the combinatorial problems.
For example,
they pre-sorted the training example for
 % the traveling salesman problem according to the counter-clockwise direction,
% or the example triangles
the Delaunay triangulation (set of triangles) by the lexicographical order 
and trained an LSTM model with the standard cross entropy \citep{vinyals15pointer}.
However, this is an ad-hoc method that depends on the particular domain knowledge
and, as we have shown, the difficulty of learning such an output was caused by the loss function that considers the ordering.
Moreover, we showed that
the standard cross entropy and the set average metrics are the less tighter upper bound of the proposed Set Cross Entropy
and also that it
empirically outperforms the standard cross entropy in the theory learning task,
even if a specific ordering is imposed on the output.
% [todo: are they using the cross entropy?]

One limitation of the current approach is 
that the set cross entropy contains a double-loop, therefore takes
$O(N^2)$ runtime for a set of $N$ objects.
However, unlike the algorithm proposed in \cite{probst2018set},
which uses a sequential Gale-Shapley algorithm which also uses $O(N^2)$ runtime,
our loss function can be efficiently implemented on GPUs because it consists of
a simple combination of logsumexp and summation.
Still, improving the runtime complexity is an important direction for future work
because the other set reconstruction tasks, including 3D point clouds datasets
like Shapenet \citep{shapenet2015}, may contain a much larger number of elements in each set.

Another direction for future work is to use the Long Short Term Memory \citep{hochreiter1997long}
for handling the sets without imposing
the shared upper bound on the number of elements in a set,
which has been already explored in the literature \citep{vinyals15pointer,vinyals2016order}.
Since our approach is agnostic to the type of the neural network, they are orthogonal to our approach.

\section{Conclusion}

In this paper, we proposed Set Cross Entropy, a measure
that models the likelihood between the sets of probability distributions.
When
the output of the neural network model can be naturally regarded as a set,
Set Cross Entropy is able to relax the search space by
making the permutations of a global minima also the global minima,
and makes the training easier.
This is in contrast to the existing approaches that try to correct the ordering
of the output by learning a permutation matrix,
or an ad-hoc methods that reorder the dataset using the domain-specific expert knowledge.

Training based on the Set Cross Entropy is robust against the dynamic dataset
which contains the vectors whose internal ordering may change time to time in an arbitrary manner.
It can also handle the sets of variable sizes by inserting dummy vectors.

We demonstrated the effectiveness of the approach by comparing Set Cross Entropy
against the normal cross entropy, as well as the other set-based metrics
such as Hausdorff distance or Set Average (Chamfer) distance.
Set Cross Entropy empirically outperformed all of these variants,
and we also provided a proof that
all of these variants are the looser upper-bound of the proposed set cross entropy.

% not necessary in journal
\fontsize{9.5pt}{10.5pt}
\selectfont

\section*{Appendix}

\subsection{Network Model for 8 Puzzle}
\label{8puzzle-model}

As mentioned in the earlier sections,
the network has a permutation invariant encoder and the fully-connected decoder.

The input to the network is a $9\times 15$ matrix, where the first dimension represents the objects
and the second dimension represents the features of each object.
The encoder has two 1D convolution layers of 1000 neurons with filter size 1, modeling the element-wise network $\rho$.
The output of these layers is then aggregated by taking the sum of the first dimension.
The result is then fed to two another fully-connected layers of width 1000,
which maps to the latent layer of 100 neurons.
All encoder layers are activated by ReLU.
The latent representation is regularized and activated by Gumbel-Softmax \cite{MaddisonMT17,jang2016categorical}
as the input is a categorical model.

The decoder consists of three fully-connected layers with dropout and batch normalization as shown below:
\begin{center}
\begin{tabular}{|l|}
 fc(1000), relu, batchnorm, dropout(0.5), \\
 fc(1000), relu, batchnorm, dropout(0.5), \\
 dense(135), reshape($9\times 15$) 
\end{tabular}
\end{center}
The last layer is then split into $9\times9$, $9\times 3$, $9\times 3$ matrices and
separately activated by softmax, reflecting the input dataset (\refig{8puzzle-features}).

\subsection{Network Model for Blockswolrd}
\label{blocks-model}

The same network as the 8-puzzle was used, except that
the input and the output is a $5\times 1224$ matrix.
The activations of the last layer is different:
The first 1024 features are activated by a sigmoid function,
while the 200 features are divided into 40, 60, 40, 60 dimensions (for
the one-hot bounding box information) and are separately activated by softmax.

\subsection{Feature Extraction for Blocksworld}
\label{8puzzle-feature}

The 32x32 RGB image patches in the Blocksworld states are compressed into the feature vectors
that are later used as the input.
The image features are learned by a convolutional autoencoder depicted in \refig{cae-encoder-implementation}
and \refig{cae-decoder-implementation}.

\begin{figure}[htb]
\centering
\begin{tabular}{|l|}
 Input($32\times 32 \times 3$),          \\
 GaussianNoise(0.1),                     \\
 conv2d(filter=16, kernel=$3\times 3$,), \\
 relu,                                   \\
 MaxPooling2d($2\times2$),               \\
 % bn,                                   \\
 conv2d(filter=16, kernel=$3\times 3$,), \\
 relu,                                   \\
 MaxPooling2d($2\times2$),               \\
 conv2d(filter=16, kernel=$3\times 3$,), \\
 sigmoid
\end{tabular}
 \caption{The implementation of the encoder for feature selection, which outputs a $8\times 8 \times 16$ tensor.}
\label{cae-encoder-implementation}
\end{figure}

\begin{figure}[htb]
\centering
\begin{tabular}{|l|}
 Input($8\times 8 \times 16$),           \\
 conv2d(filter=16, kernel=$3\times 3$,), \\
 relu,                                   \\
 UpSampling2d($2\times2$),               \\
 conv2d(filter=16, kernel=$3\times 3$,), \\
 relu,                                   \\
 UpSampling2d($2\times2$),               \\
 conv2d(filter=16, kernel=$3\times 3$,), \\
 relu,                                   \\
 fc(3072),                               \\
 sigmoid,                                \\
 reshape($32\times 32\times 3$)
\end{tabular}
 \caption{The implementation of the decoder for feature selection.}
\label{cae-decoder-implementation}
\end{figure}

\subsection{Network Model for Rule Learning}
\label{ilp-model}

We removed the encoder and the latent layer from the above models,
and connected the input directly to the decoder. While the decoder has the same types of layers, the width is shrinked to 400.
In the $n$-neighbor scenario,
the input is a $2+163(n+1)$ vector,
which consists of
a one-hot label of 2 categories for the predicate of the head,
and $n+1$ one-hot labels of 163 categories for the arguments of the head.
163 categories corresponds to the number of countries in the Countries dataset \citep{bouchard2015approximate}.
The output is a $[n, 328]$ matrix, where each row represents a binary predicate $(328 = 2 + 2\cdot 163)$.

\subsection{Example Application of the Permutation-Invariant Representation \& Reconstruction}

To address the practical utility of ``set reconstruction'' or ``set autoencoding'',
we added a new experiment. We modified Latplan
\citep{Asai2018} neural-symbolic classical planning system, a system that operates on a
discrete symbolic latent space of the real-valued inputs and runs Dijkstra's/A* search using a state-of-the-art
symbolic classical planning solver. We modified Latplan to take the set-of-object-feature-vector input rather than images.
It is a high-level task planner (unlike motion planning / actuator control) that has implications on robotic systems,
which perceives a set of inputs already preprocessed by the external system. For example, the image input is
first fed into Object Recognition system (e.g., YOLO, \citep{redmon2016you}) and the planner receives a set of feature vectors 
extracted from the image patches segmented from the raw image, rather than feeding the image input directly
 to the planning system.

Latplan system learns the binary latent space of an
arbitrary raw input (e.g., images) with a Gumbel-Softmax variational autoencoder,
learns a discrete state space from the transition examples, and runs a symbolic, systematic search
algorithm such as Dijkstra or A*  search which guarantee the optimality of the
solution.  Unlike RL-based planning systems, the search agent does not contain the learning
aspects. The discrete plan in the latent space is mapped back to the raw image
visualization of the plan execution, which requires the reconstruction capability of (V)AE.
A similar system replacing Gumbel Softmax VAE with Causal InfoGAN was later proposed \citep{kurutach2018learning}.

We replaced Latplan's Gumbel-Softmax VAE with our autoencoder used in the
8-Puzzle and the Blocksworld experiments (Appendix, \refsec{8puzzle-model},\refsec{blocks-model}). Our autoencoder
also uses Gumbel Softmax in the latent layer, but it uses \citep{zaheer2017deep} encoder
and is trained with Set Cross Entropy.

When the network learned the representation, it guarantees that the
planner finds a solution because the search algorithm 
(e.g., Dijkstra) is a complete, systematic, symbolic search algorithm,
which guarantees to find a solution whenever it is reachable in the state space.
If the network cannot learn the permutation-invariant
representation, the system cannot solve the problem and/or return the
human-comprehensive visualization. This makes the specific
permutation-invariant representation using \citep{zaheer2017deep} and
the proposed Set Cross Entropy necessary when the input is given as a
set of future vectors.

\subsubsection{8 Puzzle}

First,
we performed a training on a dataset in which the object vector ordering is randomized.
The autoencoder compresses the
$15\times 9=135$-bit binary representation (object vectors) into a
permutation-invariant 100-bit discrete latent binary representation.
We provided 5000 states for training the autoencoder, while
the search space consists of 362880 ($=9!$) states and 967680
transitions.

Note that each state have $9!$ variations due to the permutations
in the way the tiles and the locations are reported.
This also increases the number of transition quadratically ($(9!)^2$).

We generated 40 problem instances of 8-puzzle each generated by a random walk from
the goal state. 40 instances consist of 20 instances each generated by a 7-steps random walk
and another 20 by 14 steps.
We solved 40 instances using Fast
Downward classical planner \cite{Helmert04} with blind heuristics in order to remove the effect of heuristics.

We compared the number of problems successfully solved by two variations of Latplan
where each uses the autoencoder trained with Set Average and Set Cross Entropy, respectively, for encoding the input into binary latent space.
Both versions managed to solve all instances
because both Set Average and Set Cross Entropy managed to train the AE from 5000 examples with sufficient accuracy.
All solutions were correct (checked manually).
Since the search algorithm being used is optimal, the quality of the solution was also identical.

\begin{figure}[htbp]
 \centering
 \begin{minipage}[c]{0.48\linewidth}
  \centering
  \includegraphics[width=0.1\linewidth]{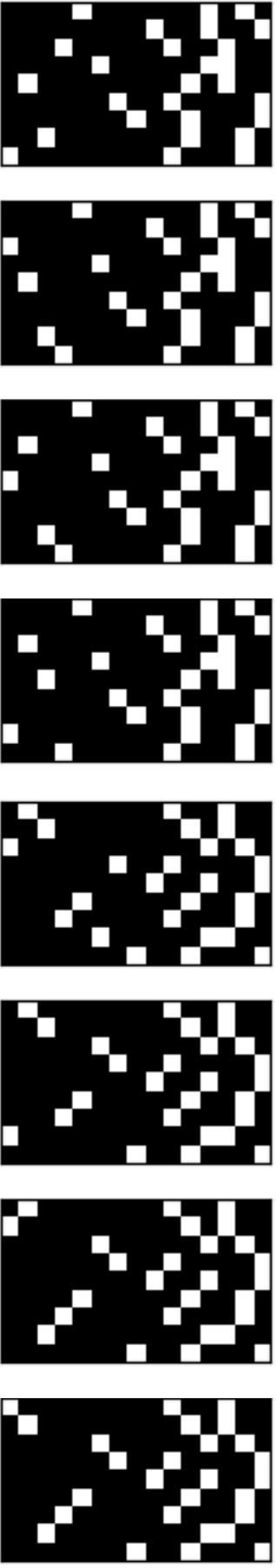}
 \end{minipage}
 \begin{minipage}[c]{0.48\linewidth}
  \includegraphics[width=\linewidth]{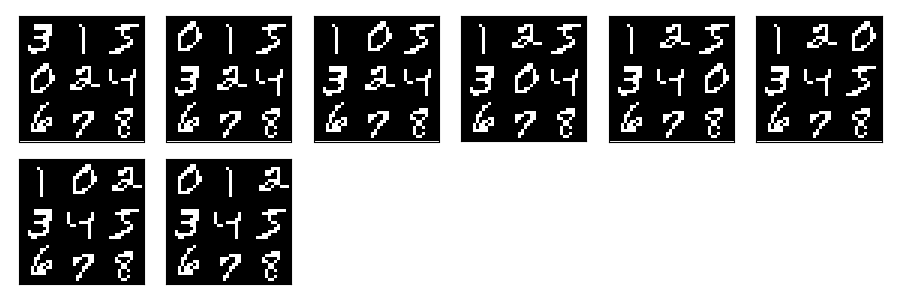}
 \end{minipage}
 \caption{
(\textbf{Left}) An example plan in a set-of-object-vector form, decoded from its binary latent representation using the permutation-invariant autoencoder (the plan is executed from top to bottom).
(\textbf{Right}) Its visualization using the tile images (taken from MNIST) pasted onto a black canvas (The plan is executed from left to right, top to bottom).
 }
 \label{8puzzle-success}
\end{figure}

\subsubsection{Blocksworld}

We solved 30 planning instances in a 4-blocks, 3-stacks environment.
The search space consists of 5760 states and 34560 transitions,
and each state has $4!$ variations due to permutations of 4 blocks.
We provided 1000 randomly selected states for training the autoencoder.
The instances are generated by
taking a random initial state and choosing a goal state by the 3, 7, or 14 steps random walks (10 instances each).
We used the same planner configuration for all instances.
The correctness of the produced plans is checked manually.

We compared the number of problems successfully solved by Latplan
between two variations of Latplan using the autoencoder trained with Set Average and Set Cross Entropy, respectively.
For the total of 30 instances, both Latplan+Set Avg and Latplan+SCE returned plans, however
the plans returned by Latplan+Set Avg were correct in 11 instances, while Latplan+SCE returned 14 correct instances (Details in \reftbl{blocksworld-planning-results}).

As the autoencoder trained by Set Average had more substantial reconstruction error, it sometimes fails to
capture the essential feature of the input, causing the system to return an invalid plan.
The common error was changing the surface of the blocks or swapping the blocks without a proper action needed,
e.g., moving more than two blocks, move a block and polish another block in a single time step.

\begin{table}[htbp]
 \centering
 \begin{tabular}{|c|cc|}
  Random walk steps & \multicolumn{2}{|c|}{The number of solved instances} \\
  used for generating & \multicolumn{2}{|c|}{(out of 10 instances each)} \\
  the problem instances & \hspace{1cm} $\mathrm{SH}$ \hspace{1cm}  & \hspace{1cm} $A_{1\mathrm{H}}$ \hspace{1cm} \\\hline
  3  & 7 & 7 \\
  7  & \textbf{5} & 3 \\
  14 & \textbf{2} & 1 \\\hline
 \end{tabular}
 \caption{The number of instances solved by Latplan using a VAE trained by Set Cross Entropy ($\mathrm{SH}$) and Set Average ($A_{1\mathrm{H}}$) of the cross entropy.}
 \label{blocksworld-planning-results}
\end{table}

\clearpage

\begin{figure}[htb]
 \centering
 \includegraphics[width=0.48\linewidth]{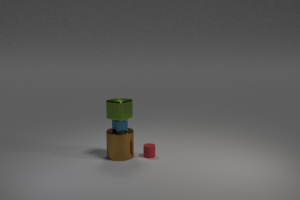}
 \includegraphics[width=0.48\linewidth]{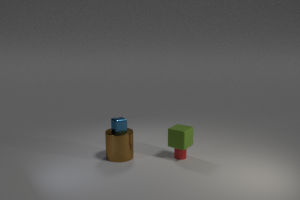}
 \caption{An example of a problem instance. (\textbf{Left}) The initial state. (\textbf{Right}) The goal state.
The planner should unpolish a green cube and move the blocks to the appropriate goal position, while also following the environment constraint that
the blocks can move or polished only when it is on top of a stack (including the floor itself).}
 \label{blocksworld-ig}
\end{figure}

\begin{figure}[htb]
 \includegraphics[width=0.24\linewidth]{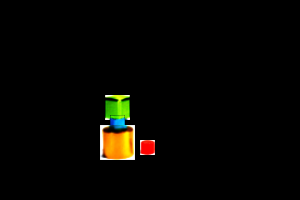}
 \includegraphics[width=0.24\linewidth]{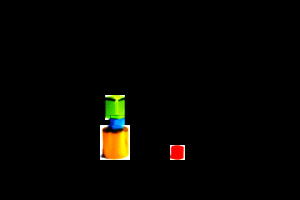}
 \includegraphics[width=0.24\linewidth]{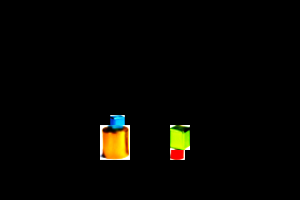}
 \includegraphics[width=0.24\linewidth]{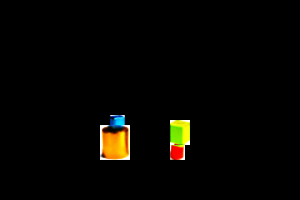}
 \caption{An example of a successful plan execution, returned by Latplan using the AE trained by the proposed Set Cross Entropy method.
The AE is used for encoding the object-vector input into a binary space that is suitable for Dijkstra search.
While the problem was generated by a 7-step random walk from the goal state, Latplan found a shorter, optimal solution because of the underlying optimal search algorithm (Dijkstra).
}
 \label{blocksworld-success}
\end{figure}

\begin{figure}[htb]
 \includegraphics[width=0.24\linewidth]{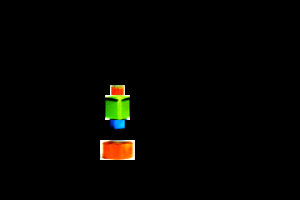}
 \includegraphics[width=0.24\linewidth]{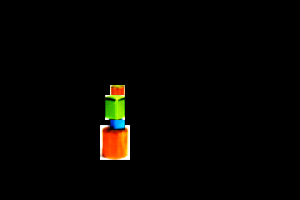}
 \includegraphics[width=0.24\linewidth]{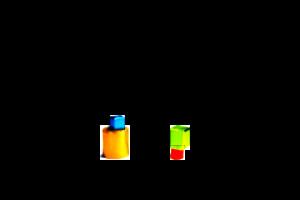}
 \includegraphics[width=0.24\linewidth]{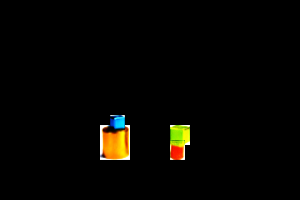}
 \caption{The decoded solution found by Latplan for the same instance, where the AE is trained by SetAverage, which had a higher mean square error for the reconstruction.
As a result, not only the initial state is invalid, but also, at the second step, two blocks are simultaneously moved in a single action, which is an invalid state transition.}
 \label{blocksworld-failure}
\end{figure}

\clearpage

\end{document}